# Enhanced Arabic-language cyberbullying detection: deep embedding and transformer (BERT) approaches

**Ebtesam Jaber Aljohani, Wael M. S. Yafooz**
Department of Computer Science, College of Computer Science and Engineering, Taibah University, Madinah, Saudi Arabia

**ABSTRACT**

Recent technological advances in smartphones and communications, including the growth of such online platforms as massive social media networks such as X (formerly known as Twitter) endangers young people and their emotional well-being by exposing them to cyberbullying, taunting, and bullying content. Most proposed approaches for automatically detecting cyberbullying have been developed around the English language, and methods for detecting Arabic-language cyberbullying are scarce. Methods for detecting Arabic-language cyberbullying are especially scarce. This paper aims to enhance the effectiveness of methods for detecting cyberbullying in Arabic-language content. We assembled a dataset of 10,662 X posts, pre-processed the data, and used the kappa tool to verify and enhance the quality of our annotations. We conducted four experiments to test numerous deep learning models for automatically detecting Arabic-language cyberbullying. We first tested a long short-term memory (LSTM) model and a bidirectional long short-term memory (Bi-LSTM) model with several experimental word embeddings. We also tested the LSTM and Bi-LSTM models with a novel pre-trained bidirectional encoder from representations (BERT) and then tested them on a different experimental models BERT again. LSTM-BERT and Bi-LSTM-BERT demonstrated a 97% accuracy. Bi-LSTM with FastText embedding word performed even better, achieving 98% accuracy. As a result, the outcomes are generalized.

***Corresponding Author:***

Ebtesam Jaber Aljohani
Department of Computer Science, College of Computer Science and Engineering, Taibah University
Madinah, Saudi Arabia
Email: eetowi@taibahu.edu.sa

## 1. INTRODUCTION

Cyberbullying is becoming increasingly prevalent in society. Although it can occur on many different online platforms and channels, cyberbullying is most prevalent on social networking sites. Bullying is a negative behavior that intentionally and repeatedly targets a particular person; bullying harms people physically and harms society. Bullying occurs in myriad cultural contexts, and bullies employ a varied vocabulary that includes terms like “disgusting,” “retarded,” and “stupid” [1], [2]. Bullying’s three main characteristics are repetition, intention to harm, and an unequal power dynamic between the victim and the offender [3] this behavior can have terrible consequences, including the death of some victims. Currently, people can digitally connect with others regardless of the distance between them. Since protecting free speech means protecting social media users from abuse, harassment, and intimidation, detecting bullying automatically is difficult [4], doing so, especially in Arabic-language posts and messages, is becoming increasingly difficult as social platforms proliferate [5].

Users of X, a popular interactive social networking platform, have recently left both positive and negative remarks. Social media users usually use pseudonyms in order to hinder efforts to monitor their behavior [6]. Previously, researchers did not regard this problem as a research concern but are now taking it seriously [7]. Numerous studies on cyberbullying have researched a variety of implementation techniques [8]–[10]. They have evaluated the diversity of machine learning implementations, which produced good results for English terms [11]–[13] and some Arabic terms [14], [15]. Others used deep learning model click or tap here to enter text [9], and these also produced good results for English terms. Some studies have combined different techniques. Further research on texts in languages other than English is required. Studies that have used Arabic texts are particularly rare. Several studies tested machine learning and other algorithms in this field, such as "decision tree (DT), Bayesian, support vector machine (SVM), random forest (RF), K-nearest neighbor (KNN)" [12], [16], on almost all English-language posts on social media. We summarized them based on three different approaches.

In the machine learning approach, multiple systems are available that can precisely and automatically detect cyberbullying. Ali *et al.* [12] sought to draw attention to earlier studies and suggest a method for identifying sarcasm in cyberbullying posts from a different datasets. They determined that the SVM classifier outperformed the classifiers stated earlier on classifying datasets 2, 3, and 4 and achieved 92% accuracy on dataset 1. Research on identifying cyberbullying is expanding because of cyberbullying's significant impact on social media users, particularly children and teenagers. While many studies have focused on identifying similar patterns in English text, not enough machine learning studies have examined Arabic-language cyberbullying. Some studies on detecting Arabic-language cyberbullying produced extremely good results while others produced average results. By using several classification methods, Kanan *et al.* [14] proposed applying machine learning to detect bad textual acts in Arabic. Various Arabic natural language processing (NLP) technologies have also been used. The results demonstrate that the F1-measure values produced by the RF algorithm were the highest. The RF algorithm's 94.1 indicated a 95% accuracy on a dataset of X posts, and SVM gave 91.7 on a dataset of Facebook posts. These outcomes held true when neither stemming nor stop-word removal was used. Other machine learning methods could improve results. Alduailaj and Belghith [15] found bullying in Arabic writings in this sector. One study examined the application of SVM and several significant data processing features to 30,000 comments. Different situations were included in it: cleaned, stemmed, segmented, and stop-word comparison using term-frequency times inverse document-frequency (TF-IDF) and bag of words (BoW). The results showed that SVM, with an accuracy of 95.742%, identified cyberbullying more accurately than naïve Bayes (NB) did. Due to our model's high accuracy, users will be better protected from social network bullies than they currently are. Other tools may increase accuracy. AlFarah *et al.* [17] used online social network data to identify Arabic cyberbullying. They created a dataset using information found on X and YouTube. They manually annotated the information to ensure the annotation was of the desired quality. They oversampled the minority class in order to address the imbalanced dataset problem. In order to compare the classifiers' performances, they used five machine learning approaches. They presented the outcomes for each performance parameter. SVM and logistic regression (LR) both achieved 88%, whereas NB had an area under curve (AUC) score of 89%. To enhance the accuracy of methods for detecting Arabic-language cyberbullying, more data should be collected, balanced, measured, and tested using deep learning models.

Several enhanced have created deep learning models based on observations from prior studies and generated highly positive results and applied to identify Arabic-language cyberbullying. Most studies have employed models created for other languages, especially English and little for Arabic. Al-Ajlan and Ykhlef [9] argued that cyberbullying's harassment and hate is a significant challenge. To bridge the knowledge gap in this field, the researchers proposed a novel convolutional neural network (CNN) algorithm for cyberbullying detection in order to eliminate the need for feature engineering and to produce prediction systems better than traditional cyberbullying detection approaches. For the algorithm, they adapted the concept of word embedding, where similar words have a similar embedding. They showed that bullying posts have similar representations—a result that could help advance detection efforts. They found that the CNN-cyberbullying algorithm, which achieved an accuracy level of 95%, outperformed conventional content-based cyberbullying detection. Al-Ajlan and Ykhlef [18] proposed using deep learning to optimize cyberbullying detection on X. Their novel approach addresses the above challenges. Unlike previous studies [9], Al-Ajlan and Ykhlef's approach represents a post as a set of word vectors rather than extracting features from posts and feeding them into classifiers. Their approach achieved good results. They can apply different deep mothed as bidirectional-long short-term memory (Bi-LSTM) may use to improve results. Ahmed *et al.* [19] trained deep learning models on Arabic text. They examined five deep classifier models: CNN, long short-term memory (LSTM), Bi-LSTM gated recurrent unit (GRU), and CNN-BiLSTM. They compiled a dataset from the website of Al Jazeera, an Arabic news organization. They utilized the TF-IDF approach to represent important terms and conducted experiments with varying layers. Of the five models, LSTM achieved the best outcome: a 92.75% success rate. The study produced some promising results, but it might have been better had Ahmed et al. used additional embedding representation vectors.

A small number of studies have trained models using hybrid techniques that combine, for example, deep and machine learning or other. Studies that use hybrid techniques stand out within a literature comprised largely of studies that used only one method. Alotaibi *et al.* [6] proposed an automatic method to detect aggressive behavior using a consolidated deep learning model with a transformer. Their technique uses multichannel deep learning based on three models' transformer blocks, a bidirectional gated recurrent unit (BiGRU), and a CNN to classify X comments into two categories: offensive (bullying) and non-offensive (non-bullying). Their novel cyberbullying detection technique achieved promising results, including an accuracy of 88%. Alotaibi *et al.*'s work could be improved by applying different deep methods with a variety of embeddings. As with English, researchers have recently attempted to increase the accuracy of methods for detecting Arabic-language cyberbullying by combining multiple approaches. There is still a need for development and additional validation in the Arabic NLP task. Hani *et al.* [16] proposed a supervised hybrid machine learning approach for detecting and preventing cyberbullying while using many classifiers to train and recognize bullying actions. They found that neural networks had the highest performance and demonstrated an F1-score of 91.9% and an accuracy of 92.8%. The SVM achieved 90.3%. The models also showed that deep neural learning models outperformed the SVM in several experiments. The study's findings demonstrated that enforced to support the learner model by obtaining contextual data from various sources. Hani *et al.* [16] also contrasted their suggested methodology with cutting-edge approaches to demonstrate that, in most instances, their strategy greatly surpassed the latter's outcomes. Bidirectional encoder from representations (BERT) is a deep attention-based language model that can find patterns in lengthy and noisy bodies of text. Rezvani and Beheshti [7] provided a contemporary machine learning technique that tunes a variant of BERT and reported an accuracy of 86% for an English-language dataset. They found that a LSTM model and BERT outperformed baseline methods on all metrics. Few studies have employed a hybrid methodology of both machine and deep learning. The Arabic content used to prepare these hybrid methods could be improved. Table 1 summarizes recent selected studies on detecting cyberbullying. We classified the studies depending on whether they used machine learning, deep learning, or another approach.

Table 1. Comparison of studies

| Study | Approaches | Class/methods | Dataset | Word representations | Language | F1/Accuracy (%) |
|---|---|---|---|---|---|---|
| [6] | Deep learning and transformer | GRU, BiGRU, CNN, RF, transformer-block | X | Word embedding | English | 88 |
| [9] | Deep learning | CNN | X | Word embedding | English | 95 |
| [16] | Machine learning and deep learning | NN, SVM | Formspring | Word embedding | English | 92.8 |
| [18] | Deep learning | CNN | X | GloVe | English | 96 |
| [20] | Deep learning | LSTM, Bi-LSTM, GRU, CNN, CNN-BiLSTM | Aljazeera.net | TF-IDF | Arabic | 92.75 |
| [21] | Deep learning | Bi-LSTM, GRU, LSTM, RNN | Facebook/X/ Instagram | - | English | 82.18 |
| [22] | Machine learning | NB, SVM | Facebook, X | - | English, Arabic | 93 |
| [23] | Deep learning | CNN, CNN-LSTM, BiLSTM-CNN | X | Word embeddings | Arabic | 81 |
| [24] | Machine learning | LR, LGBM, SGD, RF, AdaBoost (ADB), NB, SVM | X | - | English | 92.8 |
| [25] | Machine learning | NB | Arabic social media streams | - | Arabic | 96 |

The techniques of the studies summarized in Table 1, were gaps mostly designed to examine English words at the text level and thus their results are better when applied to English-language texts than when applied to texts in other languages as Arabic terms. This paper's main purpose is to present ways to enhance models for detecting Arabic-language cyberbullying on social media platforms, especially X. It also investigates deep learning on different word embeddings and a novel hybrid deep transformer. It compares it to studies on the baseline machine learning model and the transformers (BERT) model. This study makes numerous contributions to the literature on deep learning and cyberbullying detection. It builds a novel dataset of bullying and non-bullying Arabic-language posts from X. It proposes two models of deep learning. The first model comprises LSTM and Bi-LSTM deep learning models of different experimental word embeddings (i.e., TF-IDF and the modern pre-trained models Araword2vec, GloVe, FastText, and BERT).

The second model comprises hybrid deep transformers (LSTM+BERT and Bi-LSTM+BERT) used to classify Arabic-language user comments depending on whether or not they contain cyberbullying. This study evaluates the proposed models' performances on all experiments' approaches using the most commonly used metrics as accuracy, recall, and precision for classification.

The presentation in this paper is organized as follows. In section 2, the paper explains the methodology we utilized to create the proposed models, collect data, clean, preprocess data, conduct experiments, and classify the models. In section 3, it presents the results of the study, analyzes them, and then compares them together. In section 4, it presents our conclusions and potential directions for future research.

## 2. METHOD

This study's methodology involved a series of steps that began with a set of inputs and ended with the expected model outputs. Data collection and annotation—the core of the project—was the study's third phase. In the fourth stage, we processed the data and classified it using several annotators. We then applied the suggested exploration techniques and a series of algorithms to the classified data. Figure 1 shows our methodology.

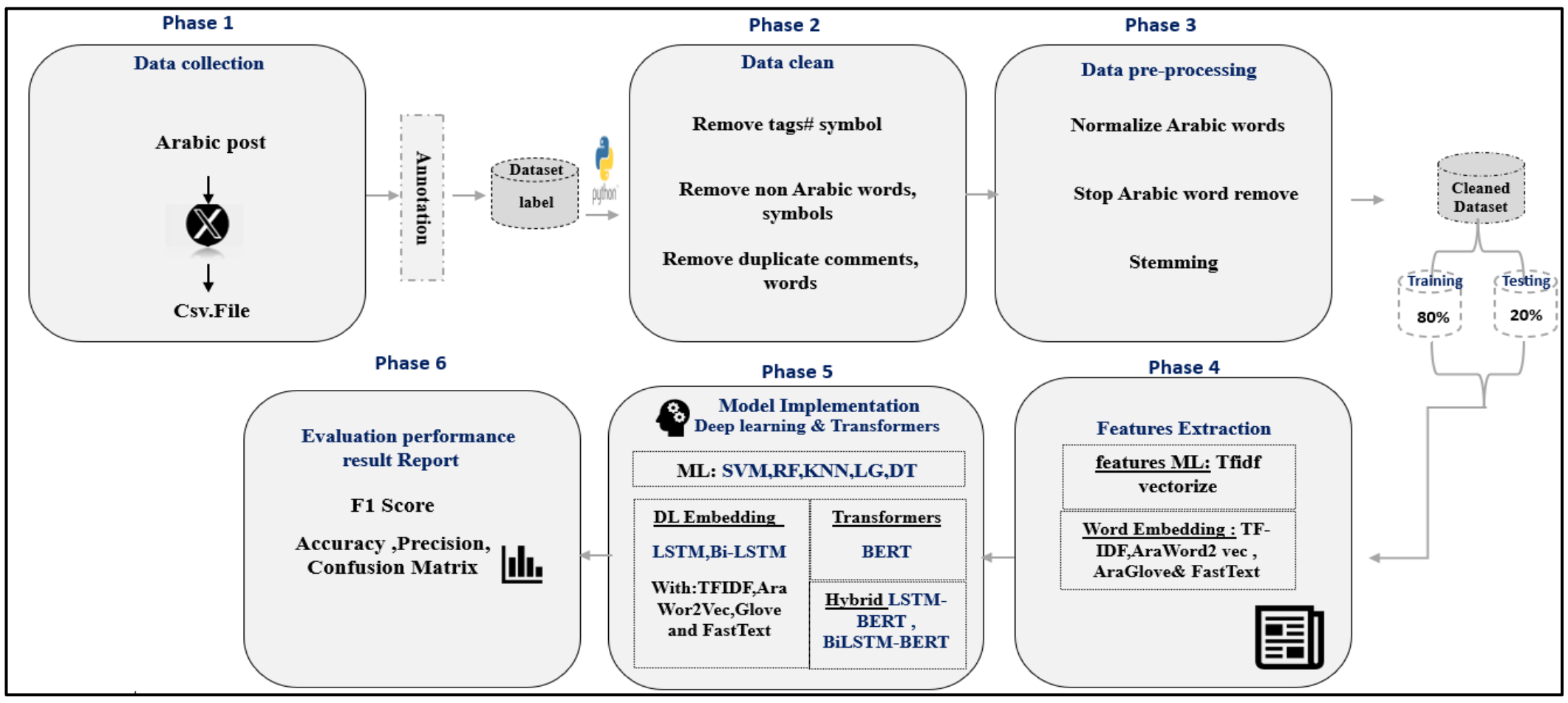

Figure 1. Methodology for studying models for detecting Arabic-language cyberbullying

### 2.1. Data collection

Since accurately testing classification methods depends on data, we had to carefully select our data source(s) and the size of our dataset. Among the obstacles that we faced were X's limits on the amount of data people are allowed to take through its API and the need to create a new account and pay for a new subscription each time we collected data. We used ntscraper and panda to collect X posts that contained terms that could indicate sarcasm or bullying (e.g., يا متخلف, يا مقرف, and غبي). Some of our key terms are listed in Table 2.

We manually classified the posts. We labeled posts that contained bullying with ones and those that did not with zeroes. Using Cohen's kappa agreement in SPSS [26], we found a kappa of 0.98, indicating good agreement between the annotators. The results of this analysis are shown in Figure 2. We saved the dataset as a CSV file and then separated the bullying and non-bullying posts. The dataset contains two fields, one for comments and the other for classification labels.

Table 2. Key terms from dataset

| Arabic term | English translation |
|---|---|
| متخلف | Ataxy-spaz |
| مقرف | Nasty |
| غبي | Stupid |

Symmetric Measures

| | | Value | Asymptotic Standard Error[a] | Approximate T[b] | Approximate Significance |
|---|---|---|---|---|---|
| Measure of Agreement | Kappa | .980 | .002 | 137.572 | <.001 |
| N of Valid Cases | | 12377 | | | |

Figure 2. Annotations' kappa agreement value

### 2.2. Data cleaning

This part shows the clean dataset methods. Cleaning and removing noise addresses X data's cluttered nature. We removed non-Arabic characters and kept only Arabic letters. We stripped out and replaced non-alphabetic characters and removed emojis, punctuation marks (e.g., #, //, !, $) and other symbols [27]. We also removed Arabic numbers and repeated characters and post rows in dataset. So, for that using some command (re.sub) in python code to remove and other marks.

### 2.3. Pre-processing

This section explains how we cleaned and pre-processed the dataset to normalize the posts' linguistic form. Pre-processing is crucial for improving the classification models' effectiveness. Table 3 contains examples of how we pre-processed and clean posts. We followed these steps:

– Normalize (text): in our dataset, we presented each Arabic character whose form can vary in a single consistent form. For example, aleph can be written in such forms as آ – إ – أ, and we normalized it to the form ا based on the recommended norm [28]. We also removed diacritics (tashdid, fatha, tanwin fath, damma, tanwin damm, kasra, tanwin kasr, sukun, and tatwil/kashida) [27].
– Remove Arabic stop words (text): we tokenized and removed stop words from Arabic text. Stop words are words that do not change how a sentence is understood. There are many stop words in Arabic, including على، الذي، التي الذين، إلخ. These words' presence in sentences expands the dataset's lexicon and makes classification more difficult [27].
– Stem (text): we used the Snowball Stemmer porter2 stemming algorithm to stem the Arabic words in our dataset. We tokenized the input text, stemmed each word, and reconstructed the text [29].

Table 3. Pre-processing and clean examples

| Original Arabic posts | After pre-processing |
|---|---|
| جميل التصوير 👏👏👏 | جميل تصوير |
| مقرف شكلهbaa | مقرف شكل |

After preprocessing and clean the dataset, it contained 10,662 Arabic-language comments: 5,736 (53.80%) bullying and 4,926 (46.20%) non-bullying. The longest post contained 1,992 words, and the shortest contained six words. We determined which terms appeared most often in the bullying and non-bullying posts. We illustrated the frequencies with which the most common words occurred by constructing word clouds with Python as shown in Figures 3 and 4.

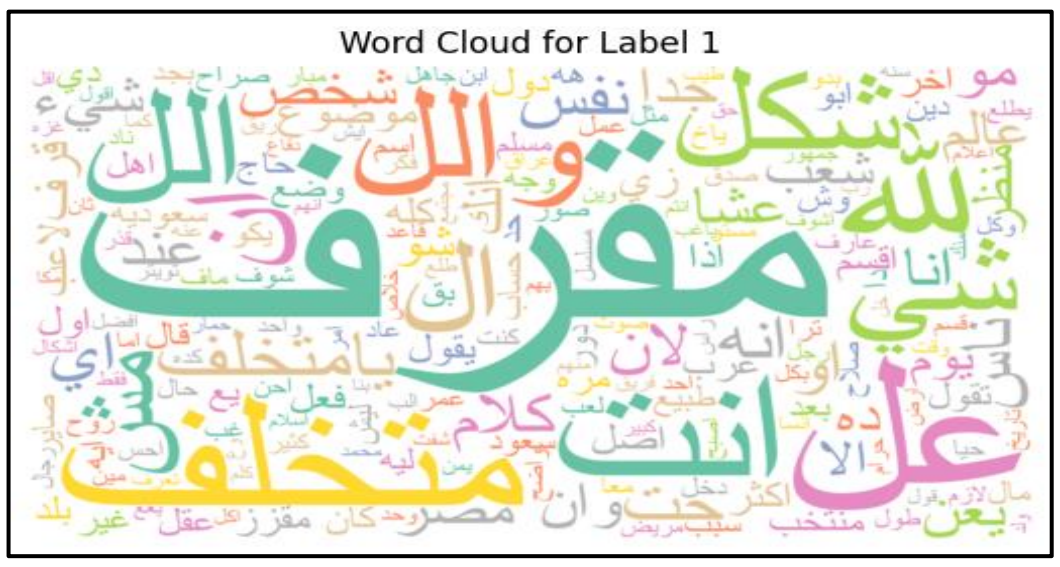

Figure 3. Most frequently used bullying words

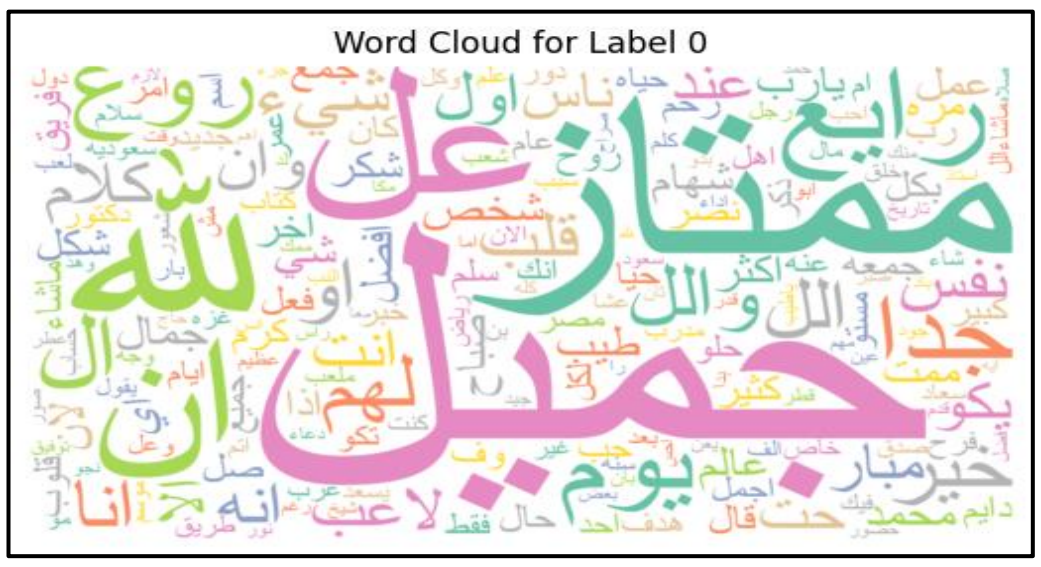

Figure 4. Most frequently used non-bullying words

### 2.4. Features extraction

We used TF-IDF, Arabic word2vec files, GloVe, and FastText techniques to convert words and phrases into low-dimensional feature vectors that automated analytical tools can use [30]. Applying neural networks, probabilistic models [31], and methods based on dimensional reduction to a word co-frequency matrix are examples of word embedding techniques. This method efficiently converts text into a numerical form that machine learning algorithms can handle and examine with ease. We used TF-IDF in the machine learning experiment and employed TF-IDF, Arabic word2vec file, GloVe, and FastText in the experiment to facilitate the use of deep learning models.

– Apply TF-IDF: we employed the TF-IDF vectorizer to transform the raw text data into numerical feature vectors. We used the scikit-learn TF-IDF weighting technique. TF-IDF is a static measurement that can be acquired as in (1) and (2) [27].

$$TF\text{-}IDF(t,d)=TF(t,d)\times IDF(t,d) \quad (1)$$

$$IDF(t)=\frac{log(n)}{df(t)}+1 \quad (2)$$

To reduce computational complexity, we limited the vocabulary size to 100 features. We converted the TF-IDF vectors to sparse NumPy arrays for further processing and constructed bigrams from the posts.

- Apply Araword2Vec: we used pre-trained Arabic word2vec embeddings to convert the text data into dense vector representations. We loaded the word2vec model from a pre-trained file (tweets_cbow_300). The type applies continues BoW. To manage memory usage, we truncated the sequences to a maximum length embedding diminution of 100. We tokenized the text data in batches and converted each word into its corresponding word2vec embedding.
- Apply GloVe: we loaded pre-trained GloVe word embeddings from a file (/content/vectors.txt) and stored them in a dictionary (embeddings_index). We created an embedding matrix based on the dataset's lexicon and the loaded GloVe embeddings.
- Apply FastText: we used pre-trained FastText to generate word embeddings for Arabic words. We downloaded and loaded the FastText model (cc.ar.300.bin). They utilized this model to convert the words in the posts into word embeddings. The preprocess_data function tokenized and padded the input comments using the FastText model and max_length_comments 90 to manage memory usage.

### 2.5. Model implementation

This section explains the models we tested in our experiments. The Arabic language has a complicated syntactic and morphological structure that makes it both challenging and semantically rich [32]. Supervised machine learning is widely used for classification. A model or learner is first trained with labeled data. The model is then tested by being made to classify sample data [33]. Deep learning algorithms have successfully used datasets that are large, dimensional, and well-organized. In this study, two deep learning models (LSTM and Bi-LSTM) with embedding and hybrid deep transformer learning classifiers in a deep neural network classified data using the varity BERT Arabic-based pre-trained model from the hugging face library. We compared these models' performances with those of baseline machine methods. The bassline models of machine learning proposed using the best-performing models we found in the relevant literature (RF, KNN, linear SVM, LR, and DT implemented in the Python working environment). By extracting features from the Arabic-language posts with TfidfVectorize() method and constructing bigrams. Also, we compare proposed models with the baseline transformer model. The following sections provide an overview of the models.

#### 2.5.1. Long short-term memory

Special kinds of recurrent neural networks (RNN) and time-RNN can process and forecast time series while avoiding issues with long-term reliance. We replaced the LSTM's buried layer neurons with a distinct set of memory cells. RNN, and its memory cell status, is crucial. The LSTM model filters information. Its door structure consists of output, input, and forgotten gates [34]. Figure 5 explains the gate structure used to update and preserve memory cell states.

#### 2.5.2. Bidirectional long short-term memory

The LSTM model's memory cells are limited to using information from the past and cannot use information from the present. Bi-LSTM provides a solution to this issue. Bi-LSTM can be thought of as two separate LSTM structures with two inputs and two hidden levels. Both of the distinct inputs employ the same sequence, but one runs forward while the other runs backward. The hidden layers connect the two inputs, and each stage's output is merged [34] as shown in Figure 6.

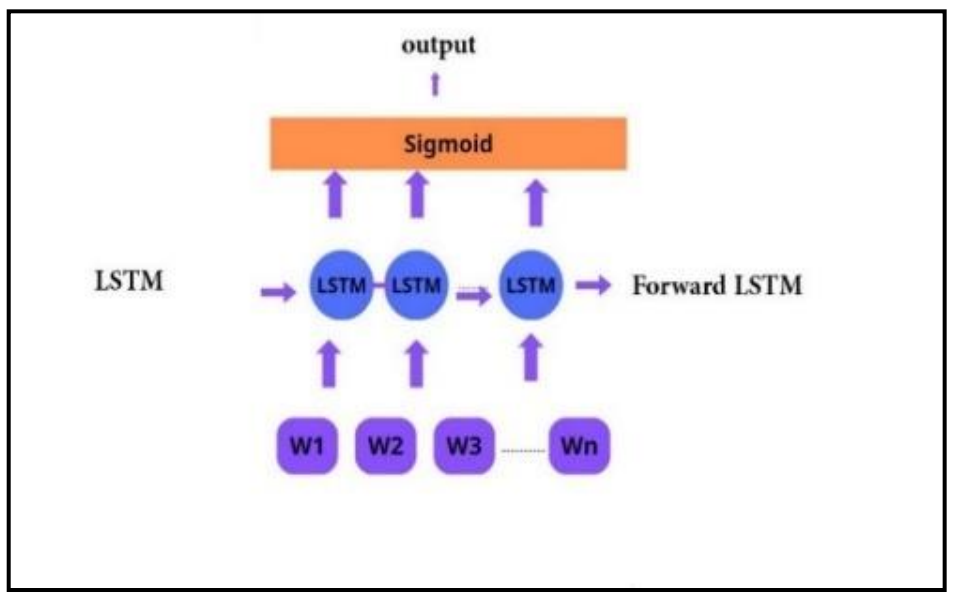

Figure 5. LSTM model

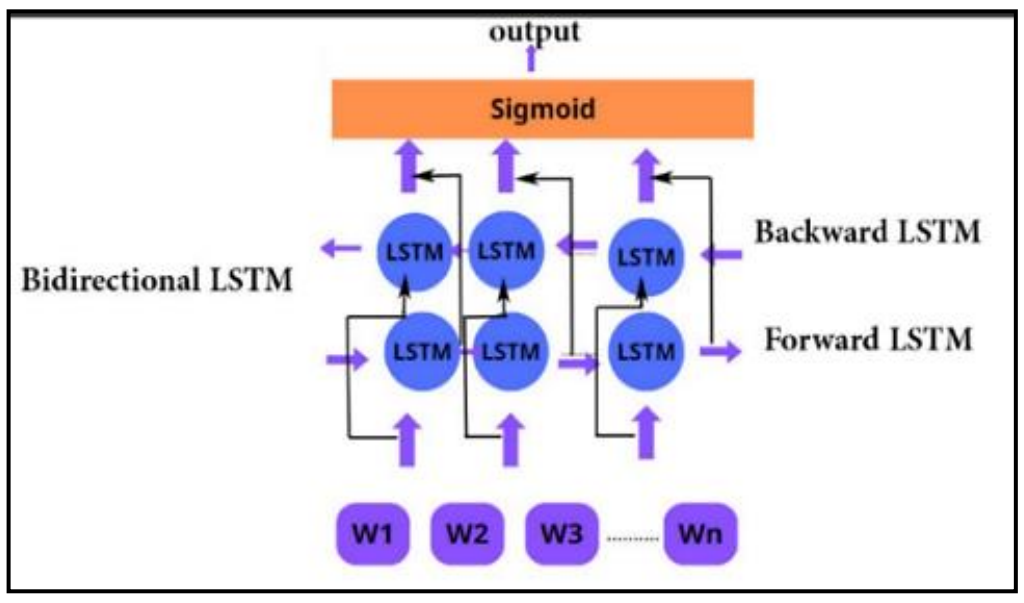

Figure 6. Bi-LSTM model

#### 2.5.3. Transformer (BERT)

BERT uses transformers architectures, an attention mechanism that recognizes contextual relationships between words (or sub-words) in a text it. BERT's multilayer bidirectional transformer encoder architecture resembles the transformer concept. BERT BASE comprises twelve levels. The encoder stack initially receives the CLS token as input before receiving a string of words as input. CLS is the categorization token. The input is then passed to the layers above. Every layer employs self-attention, transfers the outcome via a feedforward network, and then transfers control to the subsequent encoder. The model outputs a hidden-size vector (768 for BERT BASE) [35]. There are new models for Arabic dialects and posts, including aubmindlab/bert-base-Arabertv02 and aubmindlab/bert-base-large-Arabertv02. There are also other pre-trained language models, such as CAMeL for dialectal Arabic (DA), and a model pre-trained on a combination of classical Arabic, modern standard Arabic, and dialectal Arabic (Mix). These two models are known as CAMeL-Lab/bert-base-arabic-camelbert-da and Arabic camelbert -mix, respectively. Figures 7 and 8 show the structures of the BERT model and the hybrid deep transformer (BiLSTM-BERT) model.

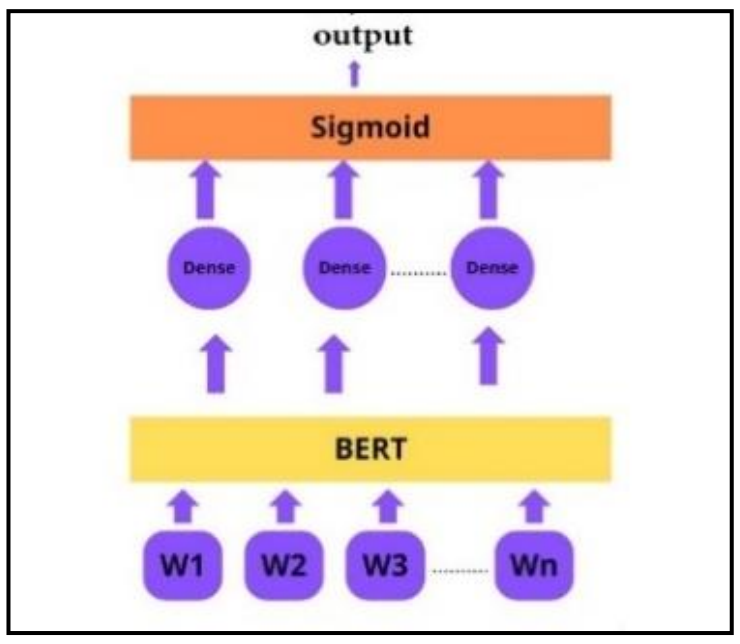

Figure 7. BERT model

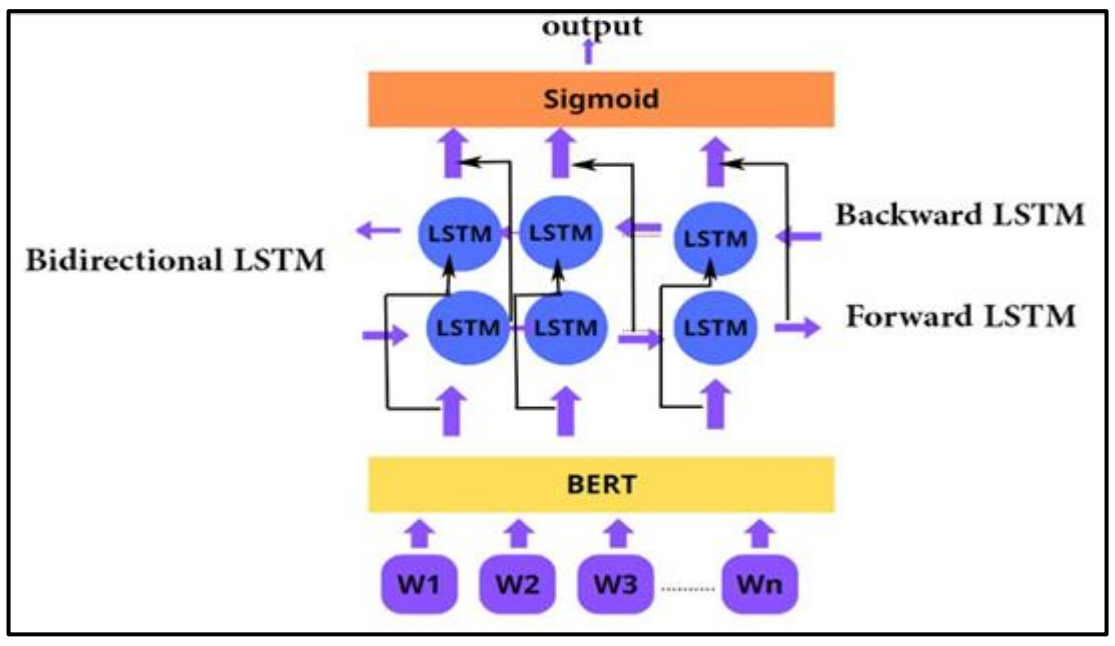

Figure 8. Hybrid deep transformer (BiLSTM-BERT) model

### 2.6. Performance evaluation

We measured the accuracy of our models and verified their effectiveness by training them on X posts and then having them filter and organize test data. We used accuracy, recall, confusion matrix, and the F1-measure to assess each classifier's output. The results of this study can easily be summarized using a confusion matrix [27], where true and false represent related and unrelated, TP and TN represent true positive and true negative, and FP and FN represent false positive and false negative. In (3), (4), and (5) are used to calculate accuracy, precision, and recall [27].

$$\text{Accuracy} = \frac{TP+TN}{TP + TN+FP+FN} \tag{3}$$

$$Precision = \frac{TP}{TP + FP} \tag{4}$$

$$Recall = \frac{TP}{TP + FN} \tag{5}$$

The F1-score [31] is calculated as the (weighted) harmonic average of precision and recall (6).

$$\text{F1} - \text{score} = 2 \times \frac{Recall \times Precision}{Precision + Recall} \tag{6}$$

## 3. RESULTS AND DISCUSSION

This section presents the experiments' settings and the experimental results of machine learning, deep learning models (deep embeddings and hybrid deep transformers), and transformer outcomes. All of the experiments were evaluated using accuracy, precision, and recall.

### 3.1. Experiment settings

This part explains our experimental setup. We used Google Colab Pro, Python 3.8, and a GPU with 50 GB of RAM. Most NLP tasks require various libraries, such as scikit-learn. Most transformer-based learning approaches have employed different BERT-based Arabic pre-trained models and hugging face's transformer library. In order to remove bias, we divided the dataset, allocating 80% for training the models and 20% for testing them. In all of the experiments, we constructed the models using Keras sequential model and compiled using the binary cross-entropy loss function and the Adam optimizer. We employed some layers containing the rectified linear unit (ReLU) activation function in certain models Bi-LSTM and LSTM. We monitored the training process and evaluated the models' performance in terms of accuracy. The following sections detail the structure of our experiments.

### 3.2. Experiments results

The result of the experiment of most common machine learning classifiers, deep learning models, transformers and proposed models have been compared in order to evaluate the performance of the proposed models. Table 4 contain the detailed results from all four experiments. We compared the models' performances.

Table 4. Comparison between the accuracy of the machine learning models, transformer, and the proposed models

| Models | Recall (%) | Precision (%) | F1-score (%) | Accuracy (%) |
|---|---|---|---|---|
| Baseline machine learning | | | | |
| SVM | 86.5 | 87.6 | 88.0 | 87 |
| LR | 82.5 | 85.4 | 82.0 | 83 |
| RF | 75.9 | 80.9 | 74.6 | 76 |
| KNN | 85.7 | 86.2 | 85.5 | 86 |
| DT | 82.4 | 85.5 | 82.1 | 82 |
| P1: Deep+word embeddings | | | | |
| LSTM-TFIDF | 98.0 | 57.4 | 72.4 | 60 |
| BLSTM-TFIDF | 97.3 | 57.0 | 71.9 | 59 |
| LSTM-GloVe | 82.6 | 83.2 | 82.9 | 82 |
| BLSTM- GloVe | 75.0 | 87.8 | 80.9 | 81 |
| LSTM-ArWor2Vec | 69.6 | 97.4 | 81.2 | 83 |
| BLSTM-ArWor2Vec | 95.4 | 98.4 | 96.9 | 97 |
| LSTM-Fasttext | 84.5 | 88.7 | 86.6 | 86 |
| Bi-LSTM- Fasttext | 96.5 | 99.7 | 98.1 | 98 |
| Transformer (BERT) | | | | |
| BERT CAMeL-da | 96.6 | 97.2 | 96.9 | 97 |
| BERT CAMeL -mix | 95.6 | 96.5 | 96.0 | 96 |
| BERT Arabertv02 | 95.4 | 95.0 | 95.2 | 95 |
| BERT-alarge-Arabertv02 | 89.2 | 95.1 | 92.1 | 92 |
| P2: Hybrid deep transformer | | | | |
| LSTM-BERT CAMeL-da | 96.3 | 97.2 | 96.8 | 97 |
| Bi-LSTM-BERT CAMeL-da | 96.8 | 97.1 | 97.0 | 97 |
| LSTM-BERT Arabertv02 | 95.6 | 95.2 | 95.4 | 95 |
| Bi-LSTM-BERTArabertv02 | 95.2 | 96.4 | 95.8 | 96 |

### 3.3. Results discussion and analysis

The results of the baseline machine learning models are shown in Table 4. TfidfVectorize() constructed bigrams from the posts for the models with certain settings to produce these results. The linear support vector classification (SVC) model performed the best. The KNN, LR, and DT models had 86%, 83%, and 82% accuracy, respectively. RF had the lowest accuracy rate (76%). The best result was obtained for the linear SVC model. These results were not superior to those of the other experiments. The results in Table 4 show that the deep learning models combined with embedding representations of the training words in Arabic posts had excellent performance. The LSTM and BI-LSTM models that used TF-IDF did not produce outstanding results. Combining the Arabic embedding Araword2Vec with Bi-LSTM achieved an excellent accuracy rate of 97%, and combining it with LSTM achieved a good result of 83%. LSTM combined with GloVe achieved 82% accuracy, and the combined Bi-LSTM-GloVe model attained 81%, which was not as good as the same model with the Araword2Vec embedding.

We combined the models with the modern embedding FastText, which differs from Araword2Vec in that it splits words into small words or n-grams. The Bi-LSTM model with FastText achieved a notably high accuracy of 98% comparing with previous study as [20], [23]. The LSTM model with FastText achieved an accuracy rate of 86%. The Bi-LSTM model outperformed of the LSTM model with both the Araword2Vec and FastText embeddings. The findings on the Bi-LTSM model with FastText are detailed in the classification report shown in Figure 9 and the confusion matrix shown in Figure 10.

In an effort to maximize the rate of model progress, we employed a transformer model pre-trained on Arabic-language X posts (Arabbertv02) and another pre-trained on a variety of dialects (CAMeL-da) in trials conducted with BERT transformer models. The model with CAMeL-da produced the best results in this experiment: 97% accuracy (comparable to the Bi-LSTM-Arwod2vec model's accuracy). In the final experiment, we merged the models that had achieved high accuracy in the earlier experiments with LSTM and with Bi-LSTM. The experiments demonstrated that the models with Bi-LSTM may perform better than others, particularly when Bi-LSTM is combined with pre-trained embeddings (i.e., transformers). Bi-LSTM-BERT produced excellent results: 97% accuracy. The LSTM and Bi-LSTM models with CAMeL-da achieved 97% inclusion and an F1-score of 96.8%. Transfer learning enhanced the LSTM model's performance such that it became comparable to that of the Bi-LSTM model with a deep learning model and embeddings. We determined the AUC scores of the best-performing models as shown in Figures 11 and 12. The findings demonstrated that:

– Bi-LSTM-FastText had the highest accuracy and F1-score, followed by Bi-LSTM-BERT, Bi-LSTM-Arword2vec, BERT CAMeL-da, and LSTM-BERT.
– Combining the transformer BERT model with CAMeL-da or Arabbert0v2 improved the LSTM model's outcomes and marginally improved Bi-LSTM's performance to 97% and accelerated its training (we stopped the hybrid Bi-LSTM-BERT CAMeL-da model's training early at 12 epoch).
– The study's demonstrated when hybrid deep learning transformers and FastText-embedding deep models can be applied they outperform on baseline machine learning models.

The drew learning curves of Bi-LSTM-FastText in Figure 13 and the hybrid transformer model in Figure 14.

Classification Report:

| | precision | recall | f1-score | support |
|---|---|---|---|---|
| 0 | 0.96 | 1.00 | 0.98 | 980 |
| 1 | 1.00 | 0.97 | 0.98 | 1153 |
| accuracy | | | 0.98 | 2133 |
| macro avg | 0.98 | 0.98 | 0.98 | 2133 |
| weighted avg | 0.98 | 0.98 | 0.98 | 2133 |

Figure 9. FastText-Bi-LSTM's classify report

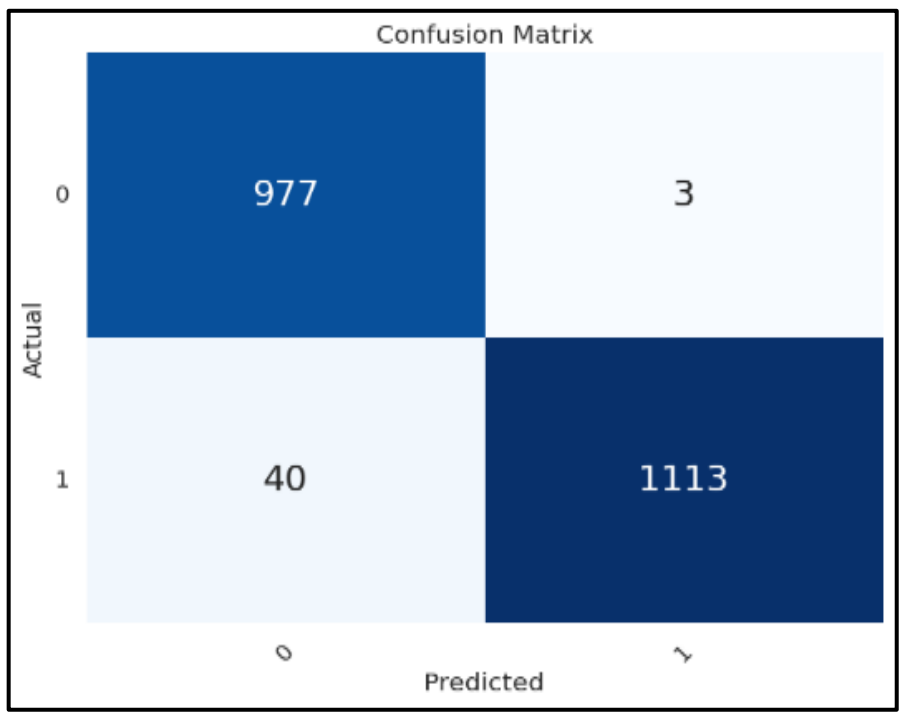

Figure 10. FastText-Bi-LSTM's

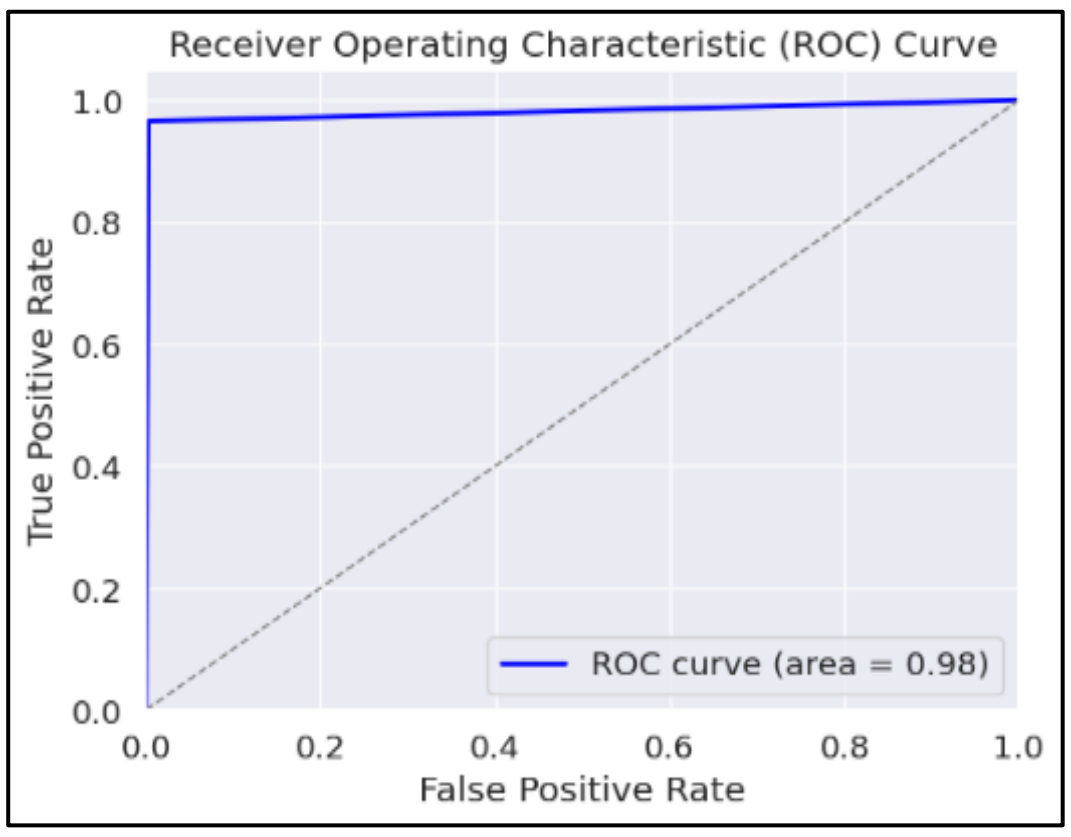

Figure 11. Bi-LSTM-FastText's AUC

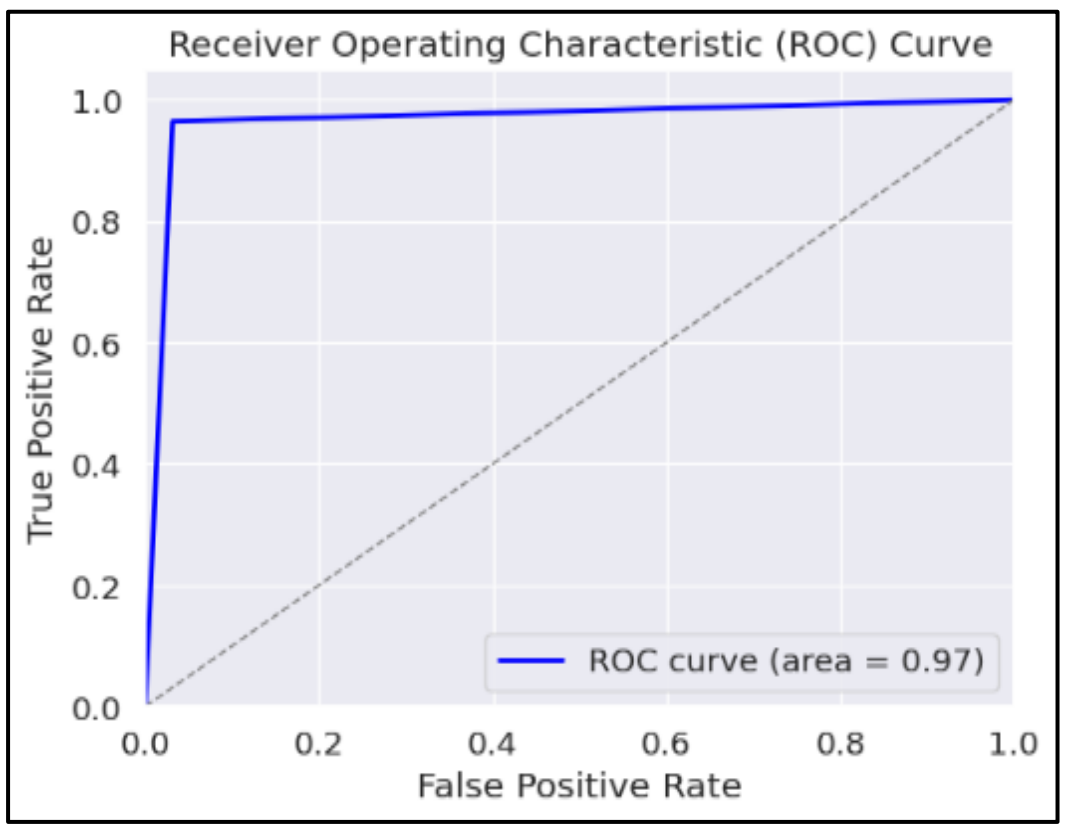

Figure 12. Hybrid transformer LSTM/Bi-LSTM-CAMeL-da's AUC

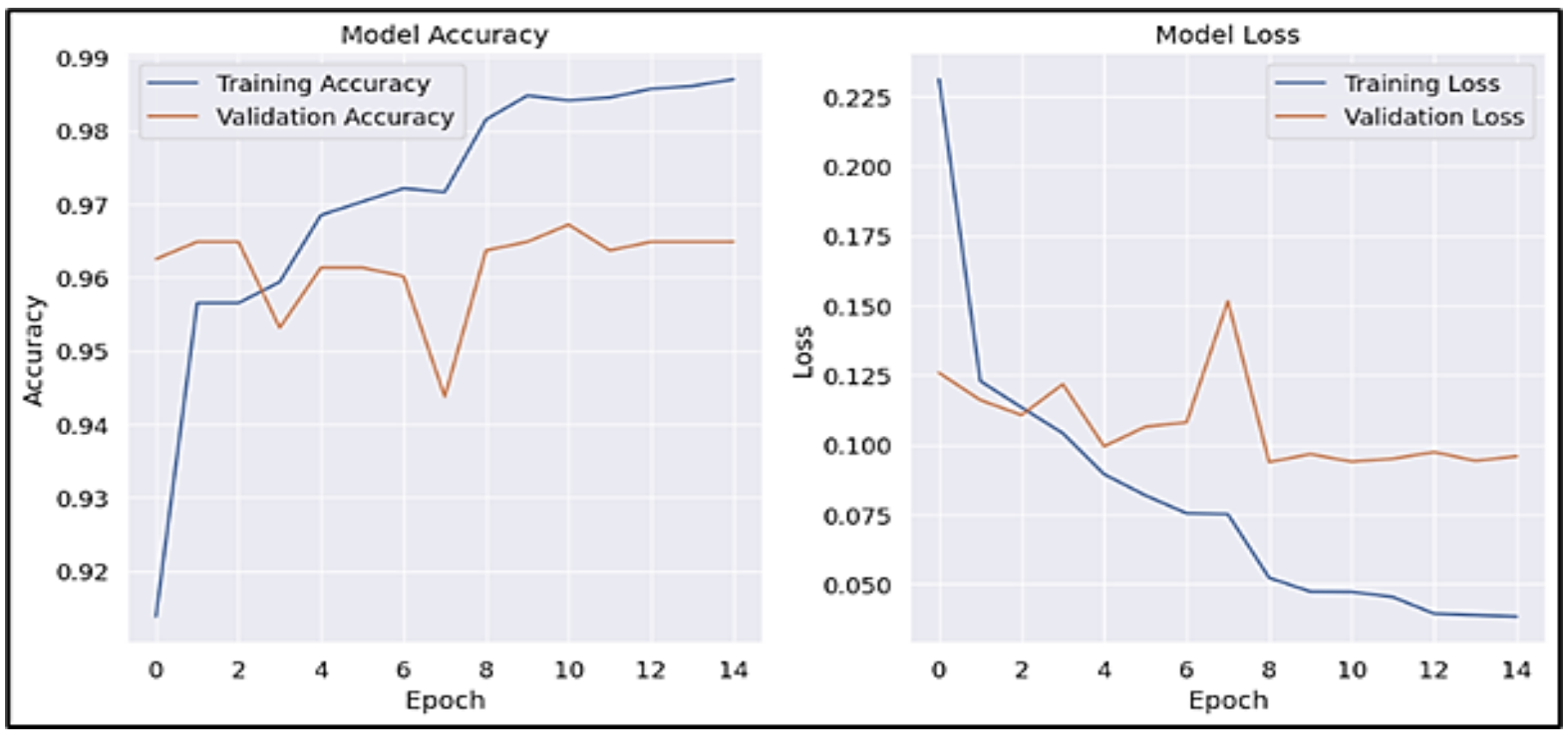

Figure 13. Bi-LSTM-FastText's learning curve

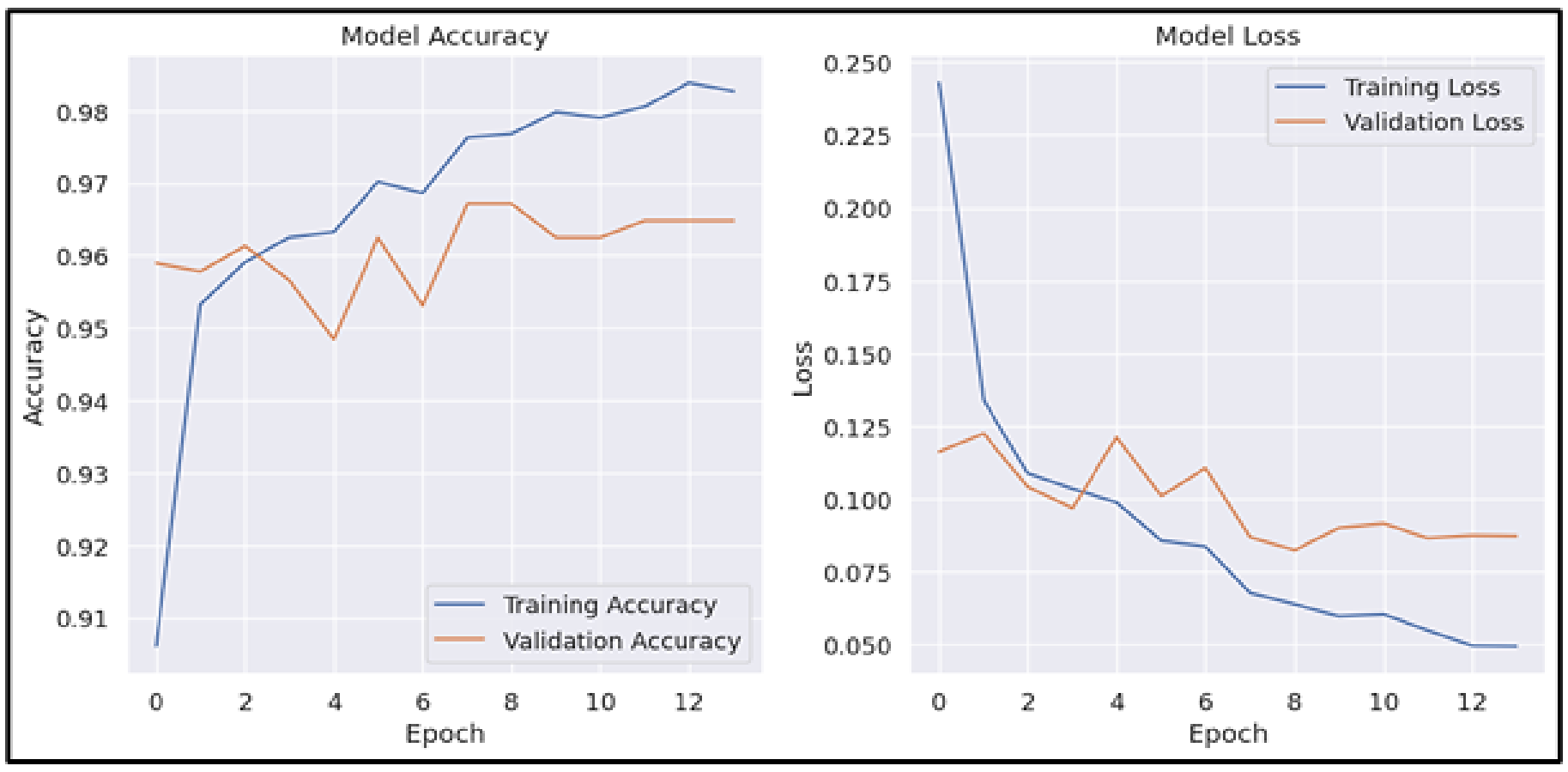

Figure 14. Hybrid transformer Bi-LSTM-CAMeL-da's learning curve

## 4. CONCLUSION

This study focuses on the detection of cyberbullying on social media, particularly on the Arabic language. By proposing two deep learning models, the first model is a Bi-LSTM-FastText model that can recognize Arabic-language bullying posts with 98% accuracy. We compared the experimental results with different baseline models to determine their accuracy rates. The second model is a combination of Bi-LSTM-BERT and LSTM-BERT models, both achieved satisfactory outcomes which is 97% accuracy when the objective was to boost the improvement. The experimental results demonstrated that the hybrid pre-trained embedding models BERT with deep learning models (LSTM, Bi-LSTM) outperformed the baseline models. In addition, the novel dataset is introduced which consists of a collection which was limited because it only applied generally to Arabic in general rather than to a specific dialect. Future research could use datasets whose size and platform of origin are different from ours to assess the models' performance on other Arabic NLP tasks, such as WhatsApp or other social media networks. Furthermore, researchers could experiment with combining various deep learning models like BiGRU with different Arabic BERT approaches to transformer techniques and assess their performance with our dataset and then integrate those combinations with transformer models. The proposed models could be deployed as tools on social media to detect Arabic bullying.

## FUNDING INFORMATION

No funding was involved.

## AUTHOR CONTRIBUTIONS STATEMENT

This journal uses the Contributor Roles Taxonomy (CRediT) to recognize individual author contributions, reduce authorship disputes, and facilitate collaboration.

| Name of Author | C | M | So | Va | Fo | I | R | D | O | E | Vi | Su | P | Fu |
|---|---|---|---|---|---|---|---|---|---|---|---|---|---|---|
| Ebtesam Jaber Aljohani | ✓ | ✓ | ✓ | ✓ | ✓ | ✓ | | ✓ | ✓ | ✓ | ✓ | | ✓ | |
| Wael M. S. Yafooz | ✓ | ✓ | | ✓ | ✓ | ✓ | | | | ✓ | ✓ | ✓ | ✓ | |

C : **C**onceptualization
M : **M**ethodology
So : **So**ftware
Va : **Va**lidation
Fo : **Fo**rmal analysis
I : **I**nvestigation
R : **R**esources
D : **D**ata Curation
O : Writing - **O**riginal Draft
E : Writing - Review & **E**diting
Vi : **Vi**sualization
Su : **Su**pervision
P : **P**roject administration
Fu : **Fu**nding acquisition

## CONFLICT OF INTEREST STATEMENT

Authors state no conflict of interest.

## DATA AVAILABILITY

The data that support the findings of this study are available on request from the corresponding author, [EB]. The data, which contain information that could compromise the privacy of research participants, are not publicly available due to certain restrictions.

## BIOGRAPHIES OF AUTHORS

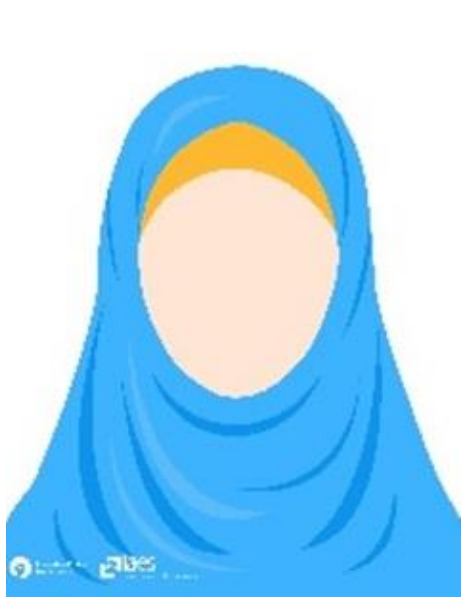

**Ebtesam Jaber Aljohani** is a lecturer in the Department of Computer Science, Taibah University, Saudi Arabia. She received a master's and B.Sc. degree in computer science. She is interested in data science, text mining, social network analytics, IoT, and artificial applications. She can be contacted at email: eetowi@taibahu.edu.sa.

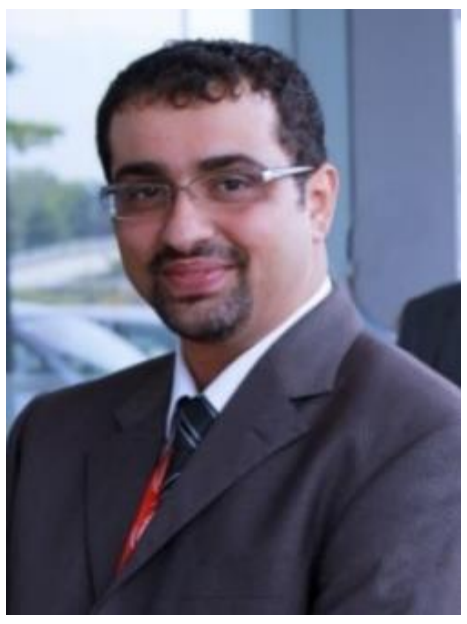

**Wael M. S. Yafooz** is an Associate Professor in the Department of Computer Science, Taibah University, Saudi Arabia. He received his bachelor degree in the area of computer science from Egypt in 2002 while a master of science in computer science from the University of MARA Technology (UiTM), Malaysia 2010 as well as a Ph.D. in computer science in 2014 from UiTM. He was awarded many Golds and Silver Medals for his contribution to a local and international expo of innovation and invention in the area of computer science. Besides, he was awarded the Excellent Research Award from UiTM. He served as a member of various committees in many international conferences. Additionally, he chaired IEEE international conferences in Malaysia and China. Besides, he is a volunteer reviewer with different peer-review journals. Moreover, he supervised number of students at the master and Ph.D. levels. Furthermore, he delivered and conducted many workshops in the research area and practical courses in data management, visualization and curriculum design in area of computer science. He was invited as a speaker in many international conferences held in Bangladesh, Thailand, India, China, and Russia. His research interest includes, data mining, machine learning, deep learning, natural language processing, social network analytics, and data management. He can be contacted at email: waelmohammed@hotmail.com or wyafooz@taibahu.edu.sa.